\documentclass[]{fairmeta}
\usepackage{amsmath}
\usepackage{tabularx}
\usepackage{booktabs}
\usepackage{placeins}
\usepackage{cuted}      
\usepackage{caption}
\usepackage{array}
\usepackage{wrapfig}
\usepackage{amssymb}
\usepackage{makecell}
\usepackage[utf8]{inputenc}
\usepackage{times}
\usepackage{latexsym}
\usepackage{xcolor}
\usepackage{colortbl}
\usepackage{enumitem}
\usepackage[T1]{fontenc}

\usepackage[utf8]{inputenc}

\newcommand{\xhdr}[1]{\vspace{1ex}\noindent\textbf{#1.}}
\title{Self-Improving VLM Judges Without Human Annotations}

\author[1,2,*]{Inna Wanyin Lin}
\author[1]{Yushi Hu}
\author[1,2]{Stella Li}
\author[2]{Scott Geng}
\author[2]{Pang Wei Koh}
\author[1,2]{Luke Zettlemoyer}
\author[2]{Tim Althoff}
\author[1]{Marjan Ghazvininejad}

\affiliation[1]{FAIR at Meta}
\affiliation[2]{University of Washington}

\contribution[*]{Work done at Meta}

\abstract{
Effective judges of Vision-Language Models (VLMs) are crucial for model development. Current methods for training VLM judges mainly rely on large-scale human preference annotations. However, such an approach is costly, and the annotations easily become obsolete as models rapidly improve.
In this work, we present a framework to self-train a VLM judge model without any human preference annotations, using only self-synthesized data. 
Our method is iterative and has three stages: 
(1) generate diverse multimodal instruction-response pairs at varying quality levels, (2) generate reasoning traces and judgments for each pair, removing the ones that do not match our expected quality levels, and (3) training on correct judge answers and their reasoning traces.
We evaluate the resulting judge on Multimodal RewardBench and VL-RewardBench across domains: correctness, preference, reasoning, safety, and visual question-answering. Our method improves a Llama-3.2-11B multimodal judge from 0.38 to 0.51 in overall accuracy on VL-RewardBench, often outperforming much larger models including Llama-3.2-90B, GPT-4o, and Claude 3.5 Sonnet, with particularly strong gains in general, hallucination, and reasoning dimensions.
The overall strength of these human-annotation-free results suggest the potential for a future self-judge that evolves alongside rapidly improving VLM capabilities.}
\correspondence{Inna Lin \email{ilin@cs.washington.edu}, Marjan Ghazvininejad 
\email{ghazvini@meta.com}} 
\date{October 2, 2025}

\begin{document}

\maketitle

\section{Introduction}

\begin{figure*}[t]
    \centering
    \includegraphics[width=\textwidth,trim={5cm 16cm 5cm 0cm}, clip]{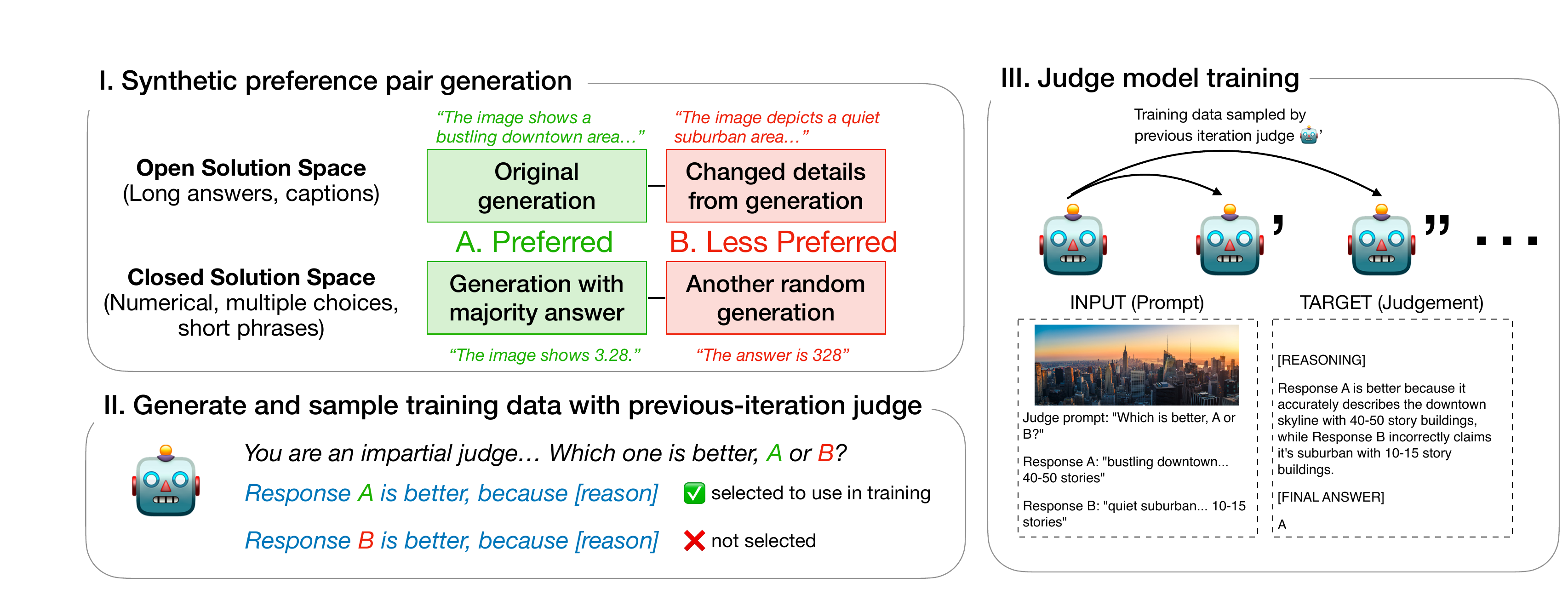}
    \caption{Self-improving VLM judge:  iterative synthetic preference data generation and judge model fine-tuning pipeline. \textbf{I. Synthetic preference pair generation.} We create synthetic preference pairs tailored to different question types. For open solution space questions (long answers, captions), we generate an original response and deliberately inject meaningful errors to create a less preferred version. For closed solution space questions (numerical, multiple choice, short phrases), we generate multiple candidates and pair the majority answer with a random alternative. \textbf{II. Iterative judge training data generation.} We use the previous-iteration judge model to evaluate the newly synthesized preference pairs and gather the judge's reasoning traces. We retain only judgments that align with our synthetic preferences. \textbf{III. Judge model training.} We fine-tune the previous-iteration judge model on these filtered reasoning traces. We iterate this three-step process several times. More details in \S\ref{sec:method}.}
\end{figure*}

VLM reward models are critical for evaluating output quality and enabling alignment with human preferences~\cite{yu2025rlaif, jing2024fgaif,2023llavarlhf, chen2024mllm}.

Existing approaches for training such model evaluators have primarily followed two directions: scaling up human preference collection and distilling from large closed-source models such as GPT and Claude (which also indirectly rely on significant human annotation)~\cite{li2024vlfeedback, llava-critic, zhang2023gpt4vevaluator, povid}. 
However, extensive human preference annotation is both costly and labor-intensive, and becomes obsolete as models advance and new tasks arise. 

In this paper, we show that it is possible to self train a VLM judge given only the VLM's own generations. 

Our method requires no human preference annotations, thereby significantly reducing the cost of reward model training. 

The key idea is that it is possible to use simple, general purpose heuristics to find or construct VLM outputs with varied quality levels, which provide enough signal to train a judge.  Our approach follows a three-step iterative process:
\begin{itemize}[noitemsep,topsep=0pt,leftmargin=*]
  \item \textbf{Synthetic preference pair generation.} We create synthetic preference pairs tailored to different question types. For questions with closed-ended answers (e.g. multiple choice or numerical problems) we generate many candidate responses and pair the majority answer with a random alternative. For open-ended questions (e.g. like captioning) we generate responses and deliberately inject meaningful errors into one version, such as changing object attributes or spatial relationships. We create substantial differences rather than minor edits, exposing the model to realistic evaluation scenarios.
  \item \textbf{Judge training data generation.} 
 
We use the previous-iteration judge model to evaluate the newly synthesized preference pairs and gather the judge's answer and reasoning traces.
Since we know the preferred pair by construction, we retain only judgments that align with our synthetic preferences.
  \item \textbf{Judge model training.} We fine-tune the previous-iteration judge model on these filtered reasoning traces and answers. 
\end{itemize}
We iterate this three-step process several times. We iteratively train a small Llama-3.2-11B model~\cite{llama3} using our framework, and evaluate the model with VL-RewardBench~\cite{vlrb} and Multimodal RewardBench~\cite{mmrb}. Our method improves the model from 0.38 to 0.51 overall accuracy, often outperforming much larger models like Llama-3.2-90B and Claude 3.5 Sonnet~\cite{claude} on VL-RewardBench.
In Section~\ref{sec:analysis}, we systematically analyze the impact of synthetic data design and iterative refinement, providing insights into the conditions under which self-improvement is most effective for VLM judge training. 
Specifically, our synthetic data construction provides a way to create a preference dataset from any visual queries, even without any reference answers.
This makes our framework applicable to new visual tasks where ground-truth annotations are unavailable or scarce, such as evaluating responses to novel image collections or emerging visual domains. Overall, our contributions include:
\begin{itemize}[noitemsep,topsep=0pt,leftmargin=*]
    \item A method to create diverse synthetic preference data for VLMs without human preference annotations, using majority voting consensus for closed-ended tasks and controlled error injection for open-ended tasks.
    
    \item An iterative self-improvement framework that trains VLM judges by filtering and learning from self-generated reasoning traces, enabling iterative refinement of evaluation capabilities.
    
    \item Empirical demonstration that our approach enables a compact 11B model to gain substantial improvements on several reward benchmarks, even surpass much larger models (90B) and proprietary systems (Claude 3.5 Sonnet) on VL-RewardBench.
    
    \item Analysis showing scaling trend over increasing iterations, and operates effectively without access to ground-truth answers, making it applicable to emerging visual domains where annotations are scarce or unavailable.
\end{itemize}

\section{Related work}
\subsection{LLM and VLM as a judge}

LLM-as-a-judge can provide scalable reward signals for RL fine-tuning, reducing dependence on expensive human annotations while enabling more efficient preference learning. In text-only domains, large language models (LLMs) serve as automatic evaluators for tasks including summarization, dialogue, and reasoning, often achieving high consistency with human annotators.Recent work has explored various training approaches for LLM-as-a-judge systems~\cite{zheng2023llmjudge,lee2023rlaif,yuan2024srlm,cui2023ultrafeedback,lambert2024tulu, whitehouse2025j1incentivizingthinkingllmasajudge, saha2025learningplanreason}.

Similarly, vision-language models (VLMs) have been deployed as multimodal judges for captioning, visual question answering (VQA), and reasoning tasks. 
Training approaches for model evaluators have primarily followed two directions: scaling human preference collection and distilling knowledge from large closed-source models such as GPT and Claude (which themselves rely heavily on human annotation)~\cite{llava-critic, wang2025unified, vlrb, mmrb}. Some works focus on building specialized judges for single dimensions like hallucination detection using synthetic data, while general-purpose judges typically require substantial amounts of preference data~\cite{jing2024fgaif, 2023llavarlhf}.

\subsection{Synthetic Data for Model Self-Improvement}
Synthetic data has emerged as a central driver of self-improvement in both LLMs and VLMs, enabling models to generate, evaluate, and refine their own outputs. Instruction-tuning pipelines bootstrap new tasks from model-generated data to expand task coverage without human labels, while iterative refinement frameworks enable models to critique and improve their own responses~\cite{Wang2022SelfInstruct,Li2024GLAN,Madaan2023SelfRefine,Shinn2023Reflexion}. Self-rewarding and self-play approaches further demonstrate that models can generate training signals to surpass supervised baselines~\cite{alemohammad2024self}. However, uncontrolled synthetic data generation can lead to degenerative feedback loops and preference leakage, underscoring the importance of careful curation~\cite{Alemohammad2023MAD,Li2025PreferenceLeakage,liu2024bestpracticeslessonslearned}.

The multimodal domain presents greater opportunities and unique challenges. While scarce and expensive multimodal annotations create stronger incentives for synthetic data generation, VLMs face distinctive issues including hallucination~\cite{sun2023aligninglargemultimodalmodels, povid, zhou2024analyzingmitigatingobjecthallucination,leng2023mitigatingobjecthallucinationslarge} and text-image misalignment~\cite{zhou2024aligningmodalitiesvisionlarge}. Most such work relies on larger models to create synthetic data. Our work extends text-domain self-improvement methods to address multimodal-specific challenges of maintaining visual grounding and factual accuracy without human annotations or larger models.
\section{Method}
\label{sec:method}

Our self-improving multimodal judge framework consists of three key stages: synthetic preference pair generation (Section~\ref{sec:2.1}), constructing judge training data with filtering (Section~\ref{sec:2.2}), and iterative training (Section~\ref{sec:2.3}). The key idea in this process is to generate diverse data that best help the model learn how to judge. 

\subsection{Synthetic Data Generation}
\label{sec:2.1}
The core challenge in training multimodal judges is obtaining high-quality preference pairs that capture the nuanced ways VLM outputs can fail. Human annotation is expensive and doesn't scale to new domains, while relying on stronger models for data generation creates dependency on external better models. We instead ask: \emph{can a model generate its own training signal by creating and recognizing flawed outputs?} While self-improvement methods have been explored in text domains, extending these approaches to multimodal settings introduces new challenges: the model must identify failures that are fundamentally visual--such as incorrectly counting objects, misinterpreting spatial relationships, or hallucinating image content.

Our approach is entirely self-contained--the same model generates and perturbs its own responses. This ensures the model learns directly from its internal knowledge rather than distilling from an external teacher. Critically, we design our synthetic data generation to create meaningful contrasts rather than superficial differences. We then know the preference direction by construction, eliminating the need to obtain ground truth labels.

We broadly categorize evaluation tasks into two types based on the open-endedness of the answers. Datasets requiring captions or longer responses have \textit{open-ended} answers that lack objectively verifiable solutions, while datasets with multiple choice, numerical answers, or short phrases have \textit{closed-ended} answers with deterministic correct responses. Formally, let $\mathcal{D} = \{(I_i, Q_i)\}_{i=1}^n$ denote a collection of image-question pairs, where $I_i$ represents the $i$-th image and $Q_i$ the corresponding question. We denote the open-ended subset as $\mathcal{D}_{nv}$ and the closed-ended subset as $\mathcal{D}_v$. 

\subsubsection{Detail Alteration for Open-Ended Task}

Data in this category are typically captions or long answers to reasoning questions. Our approach generates pairs of responses where one contains subtle inaccuracies, creating a natural preference ranking even when the absolute correctness cannot be verified.

For each image-question pair, we produce two sets of responses using our base VLM. We designate one generation as the original response and use the other to create altered versions by prompting the model to introduce specific inaccuracies, such as swapping numbers or object attributes, while preserving the original style, structure, and length. By introducing these targeted errors, we create challenging cases where the altered responses remain contextually plausible and internally coherent—mentioning objects or details that could reasonably appear in such scenes but are not actually present in the specific image. This tests a known weakness in VLMs: the tendency to conflate what is visually plausible with what is visually present~\citep{liu2024mitigatinghallucinationlargemultimodal,povid}. he judge must ground evaluations in fine-grained visual details rather than textual plausibility or language priors. A list of prompts used to alter the generations can be found in Appendix~\ref{app:alter_prompts}.

For caption tasks specifically, we aim for diverse synthetic responses to increase variety in styles and lengths. This way we try to avoid that the base model always generate captions in its own default style. We randomly vary prompts with different length specifications (short or long) and style examples. The list of prompts we used to generate in two styles for captions can be found in Appendix~\ref{app:generation_prompt_caption}. 

The underlying assumption is that while we cannot verify absolute accuracy, systematically introducing errors creates a reliable preference ranking where the original is preferred over the degraded version. Formally, for each datapoint $(I, Q) \in \mathcal{D}_{nv}$, we generate two sets of responses using our base VLM $M_{\text{base}}$. From the two generation sets, we designate one as the original response $T = M_{\text{base}}(I, Q)$ and use the other to create altered versions. We then prompt $M_{\text{base}}$ to introduce inaccuracies into these generations, producing degraded responses $T'$. This generates preference pairs $(I, Q, T, T')$ where $T$ is preferred over $T'$.

\begin{table}[h]
\centering
\small
\begin{tabularx}{\columnwidth}{p{2.5cm}|X}
\hline
\textbf{Image} & [City skyline] \\
\hline
\textbf{Original Answer (T)} & The image shows a bustling downtown area with modern skyscrapers reaching heights of approximately 40--50 stories. The glass facades reflect the afternoon sunlight. \\
\hline
\textbf{Altered Answer (T$'$)} & The image shows a quiet suburban area with colonial buildings reaching heights of approximately 10--15 stories. The brick facades absorb the evening moonlight. \\
\hline
\end{tabularx}
\caption{Example of detail alteration for synthetic data generation}
\end{table}

\subsubsection{Majority Voting for Closed-Ended Tasks}
In closed-ended answers, simply altering details in responses is insufficient—errors in logical reasoning cannot be effectively simulated by changing descriptive information. Moreover, for short answers without reasoning traces, directly changing the answer (e.g., as a simple example, from ``The answer is A'' to ``The answer is D'') does not provide meaningful learning signal, as the resulting pairs can be trivially easy when the wrong answer is obviously incorrect. Therefore, we leverage the closed-ended nature of these answers by using majority voting to identify likely correct responses, which we then pair with randomly sampled alternatives as likely incorrect responses.

For closed-ended answers $(I, Q) \in \mathcal{D}_v$, we generate $N$ responses from the base model. We select $T^*$ as the response chosen by the majority of the $N$ generations, and randomly sample a different answer $T^- \neq T^*$ as the less preferred response, creating preference pairs $(I, Q, T^*, T^-)$. To ensure the quality of the majority answer, we only include examples where at least 5 responses are identical, filtering out cases with greater diversity.

\subsubsection{Correct Answer Filtering}
As an alternative approach to majority voting, we also explore using ground truth labels from the original datasets to construct preference pairs. Note that these labels represent the correct answer to the question rather than human preferences over response quality. This provides an interesting point of comparison: while majority voting derives supervision from the model's own consistency, gold label filtering relies on external ground truth that may not always align with what the model can reliably distinguish.

For each $(I, Q)$ with gold label $G$, we generate $N$ responses and construct preference pairs $(I, Q, T^+, T^-)$ where $T^+$ matches the gold standard $G$ and $T^-$ does not.

\subsection{Training Data Generation}
\label{sec:2.2}
We sample the model's own reasoning traces where judgments align with our synthetic preference assumptions, inspired by self-training literature~\cite{wang2024selftaughtevaluators, hu2024visual}. For each preference pair $(I, Q, T^+, T^-)$, we query the current judge $M_{\text{judge}}^{(k)}$ at iteration $k$ to generate $N$ judgments:
\[
\{J_1, J_2, \ldots, J_N\} \;=\; \{\,M_{\text{judge}}^{(k)}(I, Q, T^+, T^-)\,\}_{j=1}^N
\]

Each judgment $J_j$ contains a reasoning trace $R_j$ and a binary decision $D_j \in \{0, 1\}$ indicating whether $T^+$ is preferred. We construct our training set $\mathcal{T}^{(k+1)}$ by selecting only samples where the model correctly identifies the preference:
\[
\mathcal{T}^{(k+1)} \;=\; \{(I, Q, T^+, T^-, R_j, 1) \;:\; D_j = 1\}
\]

To mitigate positional bias--the tendency for models to favor the first response position--we conduct this process with both orderings: $(T^+, T^-)$ (where the preferred response appears first) and $(T^-, T^+)$ (swapped order), retaining only pairs where the model makes the correct judgment in both configurations.  Correct judgments in both orderings indicate that the model's reasoning signal is stronger than its positional bias, reducing cases where the model arrives at the right answer for the wrong reasons. 

Intuitively, we only ``listen'' to the model's reasoning when it gets the answer right, bootstrapping from its own reasoning without amplifying mistakes. This focuses learning on reasoning patterns that consistently lead to correct judgments. 

\subsection{Training}
\label{sec:2.3}
We use supervised fine-tuning to train the judge using the curated dataset $\mathcal{T}^{(k+1)}$. The training objective maximizes the likelihood of generating correct reasoning traces and judgments:
\begin{align*}
\mathcal{L} 
  &= - \sum_{(I, Q, T^+, T^-, R, D) \in \mathcal{T}^{(k+1)}} 
       \log P(R, D \mid  I, Q, T^+, T^-)
\end{align*}
The input format combines the image, question, and candidate responses, while the output includes both the reasoning trace and final judgment.

\section{Experimental Setups}
\subsection{Dataset}
We construct our synthetic training set using the LLaVA-OneVision~\cite{li2024llavaonevisioneasyvisualtask}, which covers diverse multimodal tasks including reasoning, math and coding, and captioning. To focus on multimodal evaluation, we restrict ourselves to the single-image subsets.
To avoid dominance by any single sub-dataset, we impose a cap of 10k on the number of examples drawn from each sub-dataset, resulting in 100k prompts. This ensures balanced coverage across categories while maintaining manageable data size.

\begin{table*}[t]
\centering
\small
\setlength{\tabcolsep}{6pt}
\begin{tabular}{lcccccccccccc}
\toprule
 & \multicolumn{4}{c}{\textit{VLRB}} & \multicolumn{6}{c}{\textit{MMRB}} \\
\cmidrule(lr){2-5} \cmidrule(lr){6-11}
\textbf{Model} & \cellcolor{gray!20}\textbf{Ave.} & \textbf{Gen.} & \textbf{Hallu.} & \textbf{Reason.} & \cellcolor{gray!20}\textbf{Ave.} & \textbf{Gen.} & \textbf{Know.} & \textbf{Reason.} & \textbf{Safe.} & \textbf{VQA} \\
\midrule
\multicolumn{11}{l}{\textit{Larger Models}} \\
Claude-3.5 & \cellcolor{gray!20}0.536 & 0.434 & 0.550 & 0.623 & \cellcolor{gray!20}0.721 & 0.652 & 0.739 & 0.669 & 0.687 & 0.856 \\
GPT-4o & \cellcolor{gray!20}0.624 & 0.491 & 0.676 & 0.705 & \cellcolor{gray!20}0.713 & 0.658 & 0.720 & 0.649 & 0.668 & 0.872 \\
Llama-90B & \cellcolor{gray!20}0.539 & 0.426 & 0.573 & 0.617 & \cellcolor{gray!20}0.618 & 0.642 & 0.612 & 0.547 & 0.519 & 0.771 \\
\midrule
\multicolumn{11}{l}{\textit{Our Method (Based on Llama-3.2-11B-Vision-Instruct)}} \\
Base & \cellcolor{gray!20}0.383 & 0.297 & 0.304 & 0.549 & \cellcolor{gray!20}0.499 & 0.590 & 0.506 & 0.517 & 0.317 & 0.565 \\
Iteration 1 & \cellcolor{gray!20}0.452 & 0.398 & 0.362 & 0.596 & \cellcolor{gray!20}0.519 & \textbf{0.625} & 0.532 & 0.500 & 0.289 & 0.651 \\
Iteration 2 & \cellcolor{gray!20}\underline{0.521} & \underline{0.453} & \textbf{0.529} & 0.580 & \cellcolor{gray!20}0.521 & 0.591 & 0.506 & 0.513 & 0.304 & \underline{0.693} \\
Iteration 3 & \cellcolor{gray!20}0.488 & 0.425 & 0.426 & \textbf{0.612} & \cellcolor{gray!20}\underline{0.538} & \underline{0.599} & \underline{0.540} & \textbf{0.543} & 0.307 & \textbf{0.701} \\
Iteration 4 & \cellcolor{gray!20}\textbf{0.538} & \textbf{0.503} & \underline{0.514} & \underline{0.596} & \cellcolor{gray!20}\textbf{0.539} & 0.591 & \textbf{0.556} & \underline{0.531} & \textbf{0.329} & 0.689 \\
\bottomrule
\end{tabular}
\caption{Performance comparison on VLRB and MMRB benchmarks. Our iteratively trained judge model based on Llama-3.2-11B achieves competitive performance with significantly larger models across multiple evaluation dimensions.}
\label{tab:vlrb_mmrb_results}
\end{table*}

\subsection{Evaluation}

We evaluate our VLM judge and baselines on two comprehensive benchmarks: Multimodal RewardBench~\cite{mmrb} and VL-RewardBench~\cite{vlrb}. Both benchmarks measure judge performance across multiple dimensions including general instructions, hallucinations, correctness, reasoning, and VQA. Multimodal RewardBench comprises 5,211 preference pairs covering both long-form and short-answer formats, with labels derived from ground truth correctness or high-quality human expert annotations. VL-RewardBench spans general multimodal instructions, hallucination detection, and complex reasoning tasks, combining human-verified and AI-annotated preference labels to ensure comprehensive coverage of real-world scenarios.

\subsection{Implementation details}

We fine-tune the Llama-3.2-11B-Vision-Instruct model for 5 epochs using a learning rate of 1e-5 and a batch size of 2 per GPU across 8 GPUs (effective batch size of 16). We employ FSDP for distributed training and enable fast kernels for computational efficiency. We use cross-entropy loss computed over the generated reasoning traces from each iteration cycle. 
We iterate the training process until overall performance plateaus, operationally 
defined as less than 1\% relative improvement over three consecutive iterations on either of the benchmarks. 
In our experiments, we observe diminishing returns after iteration 4, though individual dimensions may peak at different iterations due to varying learning 
dynamics across task types. More details on the hyperparameters in data generation, sampling, and training can be found in Appendix~\ref{app:hyperparameters}.

\section{Results}

\subsection{Iterative Training Improvements} Table~\ref{tab:vlrb_mmrb_results} demonstrates the effectiveness of iterative training on our Llama3-11B based judge model. On VLRB, our model improves from 0.383 (base) to 0.538 (iteration 4), representing a 40.5\% relative gain. On MMRB, we observe improvement from 0.499 (base) to 0.539 (iteration 4), a 7.5\% relative gain. 
The progression across iterations demonstrates relatively consistent improvement, though the improvement trajectory can vary across different dimensions.

Compared with larger models, despite using only 11B parameters, our model achieves competitive performance on several dimensions: on general instruction following, our model (0.503) substantially outperforms the 90B parameter Llama-3.2-90B (0.426), Claude-3.5-Sonnet (0.434) and GPT-4o (0.491). Across VLRB hallucination and MMRB VQA, our model (0.514, 0.689) approaches larger models despite the base model's significantly weaker starting performance (0.304, 0.565).

\subsection{Performance Gains Across Dimensions}

Our iterative training exhibits dimension-specific learning dynamics that reveal the differential effectiveness of our synthetic data generation strategies. On VLRB, general instruction following demonstrates the largest relative gain of 69\% (0.297 $\rightarrow$ 0.503), indicating that our preference pair construction successfully captures the supervision signal necessary for open-ended multimodal instruction assessment. Hallucination detection on VLRB achieves 40.9\% relative improvement (0.304 $\rightarrow$ 0.514), validating our detail alteration methodology for training factual consistency discrimination. Reasoning evaluation shows 8.6\% relative improvement (0.549 $\rightarrow$ 0.596 at iteration 4), though exhibits non-monotonic optimization behavior with peak performance at iteration 3 (0.612), suggesting potential signs of overfitting in later iterations.

On MMRB, VQA tasks yield substantial gains (18.0\% relative improvement, 0.565 $\rightarrow$ 0.689), confirming effective transfer from our majority voting-based preference construction for closed-ended answers. In contrast, several dimensions show relatively marginal improvement: reasoning capability increases from 0.517 to 0.531 (2.7\% relative gain), general evaluation (averaging correctness and preference) stayed flat from 0.590 $\rightarrow$ 0.591, and safety evaluation demonstrates limited responsiveness (0.317 $\rightarrow$ 0.329). These marginal gains indicate potential distributional mismatch between our synthetic data and the supervision required for these specific capabilities.

\section{Analysis and Discussion}
\label{sec:analysis}
Our results show that self-improvement provides a promising path for building multimodal judges without relying on costly human annotations or stronger teacher models. In this section, we reflect on the strengths and limitations of our approach.  

\subsection{Scaling number of iterations}
\begin{figure}[h]
  \centering
  \includegraphics[width=0.70\columnwidth]{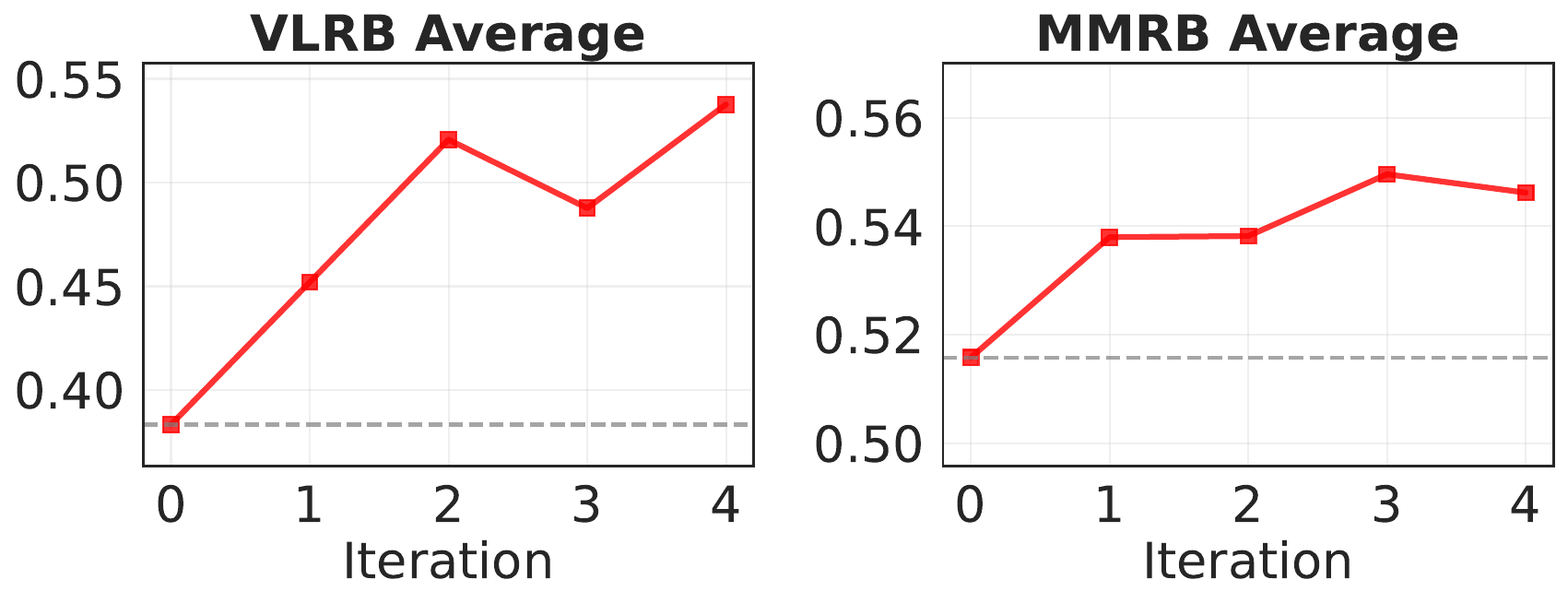}
  \caption{Judge model performance across training iterations. The left lanel shows average VLRB scores and the right panel shows average MMRB scores. After 4 iterations, our 11B judge model is comparable with Claude-3.5 and Llama-90B on VLRB.}
  \label{fig:results_ave}
\end{figure}
 
Our iterative training scheme enables the model to \textit{generate more training data pairs at each iteration} as the judge model's performance improves. \begin{wrapfigure}{r}{0.25\textwidth}
  \includegraphics[width=0.25\textwidth]{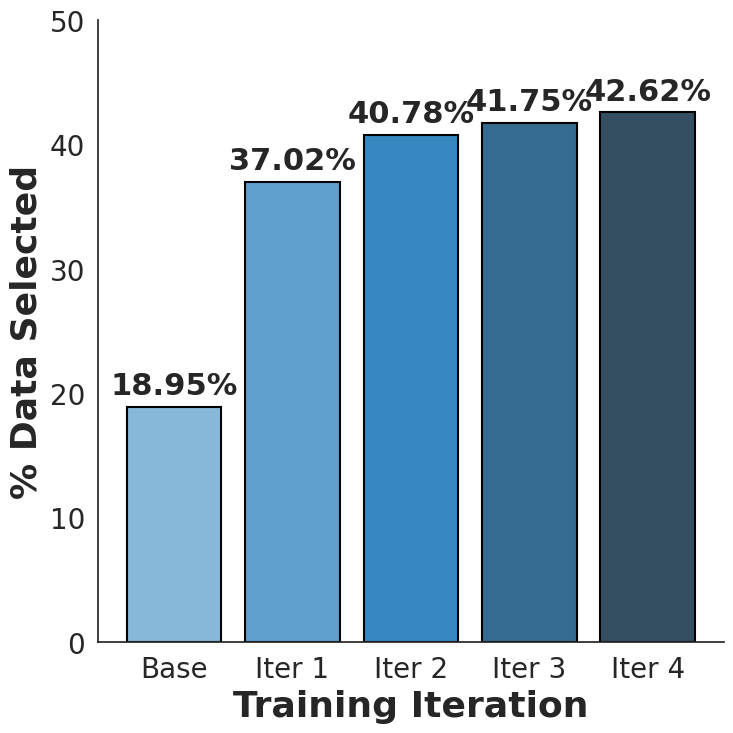}
  \caption{Increasing \% of data sampled from each training iteration}
  \label{fig:data_select_perc}
\end{wrapfigure}
Figure~\ref{fig:data_select_perc} shows that the percentage of retained training data increases from 19\% at baseline to 43\% by iteration 4, with the largest gains in early iterations.

Beyond increasing the number of training examples, we hypothesize that \textit{the quality of reasoning also improves across iterations}. Blind manual evaluation of 20 randomly sampled examples showed later iterations 
produced superior reasoning in 55\% of cases versus 10\% for early iterations, 
confirming genuine quality improvement beyond metrics. An demonstrative example of the reasoning progression for the same pairs across training iterations are shown in Appendix~\ref{tab:reasoning_progression}.

Iterative training yields substantial improvements, with the largest gains in the first iteration (10\% average improvement across VLRB and MMRB) and continued progress in later iterations (3\% average per iteration). However, improvement rates differ across dimensions. For instance, VLRB General and Hallucination show rapid, monotonic gains, while MMRB General Correctness exhibits fluctuations. These patterns reveal that different task domains respond to iterative refinement at different rates: some capabilities converge quickly and steadily with each training cycle, while others exhibit slower or more volatile learning dynamics.

\subsection{Analysis on reasoning / VQA: filtering with majority votes v.s.correct answers}
\label{sec:analysis_reasoning}

\begin{figure}[h]
  \centering
  \includegraphics[width=0.70\columnwidth]{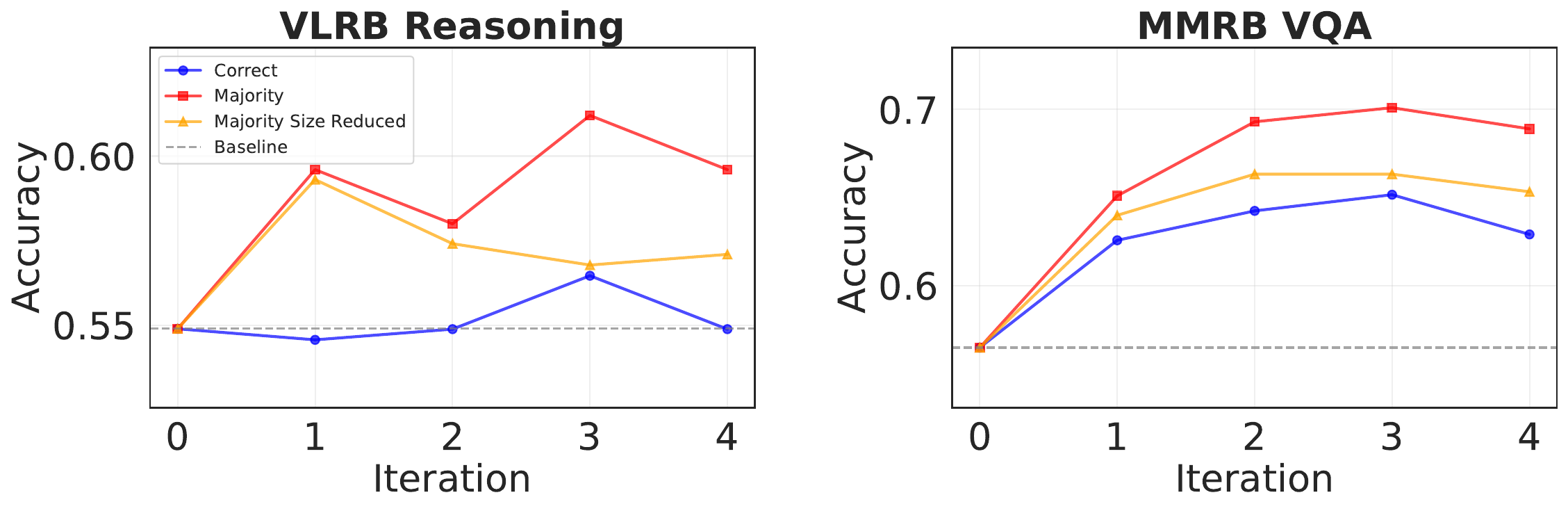}
  \caption{Performance comparing using majority voting and correct answer to filter synthetic pairs before sampling. For VLRB reasoning and MMRB VQA using majority voting to filter the synthetic pairs yields better performance after a few iterations. When reducing data size to the same with correct answer filtering, majority performance also reduced.}
  \label{fig:majority_comparison}
\end{figure}

In our preference pair synthesis, for reasoning / VQA tasks that have a short answer, we sample 16 times, and pick the majority answer as the ``preferred'' answer, and randomly sample an answer that is different as the ``less preferred'' answer. 

For ablation analysis, we also experiment with using the oracle correct answers (coming from datasets) to select the ``preferred'' answer and the ``less preferred'' answer.
Note that in the majority-voting setup, the majority answer may not be correct, especially in areas when the model struggles. 

In figure~\ref{fig:majority_comparison}, we show the comparison on dimensions where closed-ended type questions are more relevant (i.e. multiple choices, numerical answers, short answers, not captions or long answers).  On VLRB-Reasoning and MMRB-VQA, synthetic pairs constructed via majority voting yield 8.6\% and 9.5\% average improvement over correctness-filtered pairs.

While this result may appear counterintuitive, the following underlying mechanism explains our observations. First, we find that the judge model consistently samples a greater number of data points per iteration when using majority voting compared to correctness-filtered approaches.
This increased sampling capacity provides a richer training signal during each optimization cycle. As Figure \ref{fig:majority_comparison} shows, constraining the data size of the majority set to match the correct set in each iteration reduces the performance of the original majority set. However, there still exists a gap between the performance trained with the majority set and with the correct set, which leads to our second analysis. We hypothesize that even when the judge model selects the pair containing the correct answer, this may not necessarily ensure that the judge's reasoning process is sound. For instance, in the example shown in Appendix~\ref{tab:counter_example}, the judge's reasoning prioritizes stylistic factors over factual accuracy, failing to identify that Response A contains a transcription error. Such patterns suggest that the correctness of the preferred answer does not guarantee the validity of the judgment reasoning.

While using ground truth as a filtering strategy improves judge performance in this setup, majority voting filters samples more efficiently and achieves even greater performance gains. Importantly, majority voting eliminates the dependency on ground truth labels, enabling the approach to scale to any image dataset by generating questions without requiring pre-existing answers.

\subsection{Analysis on other task dimensions}
In \S\ref{sec:analysis_reasoning}, we shows the effectiveness of our method on reasoning and VQA tasks. In this section, we analyze the performance on other task dimensions. Specifically, we focus on the categories that exhibit the most and the least significant improvements.
\label{sec:analysis_bytask}
\begin{figure}[h]
  \centering
  \includegraphics[width=0.60\columnwidth]{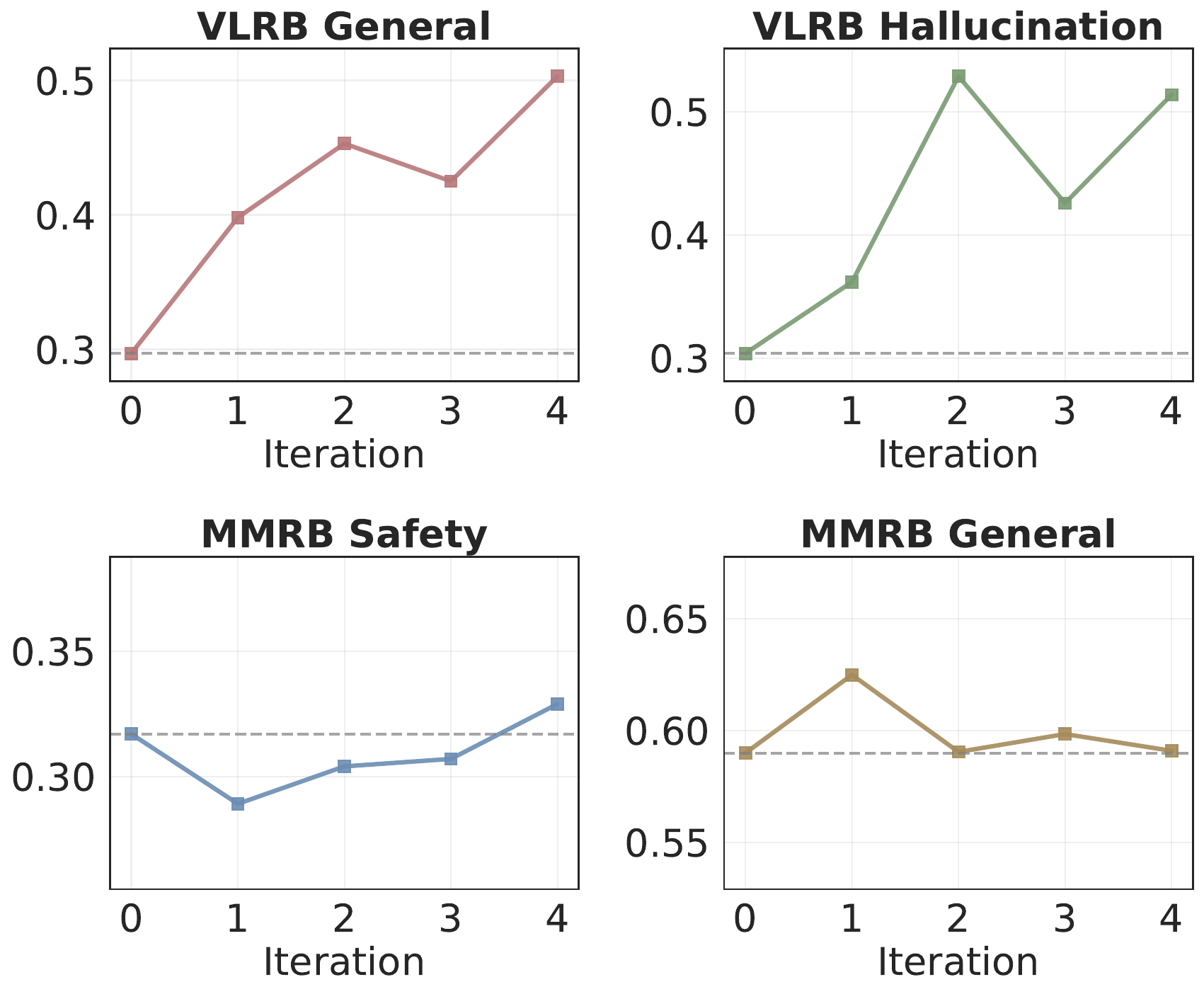}
  \caption{The dimensions that showed the most significant improvements (VLRB General, Hallucination) and the least significant improvements (MMRB Safety, General). More details in \S\ref{sec:analysis_bytask}.}
  \label{fig:your_label}
\end{figure}

\xhdr{VLRB General \& Hallucination} Our self-improving pipeline demonstrates particularly strong performance on VLRB general and hallucination categories. On VLRB general dimensions, our iteratively trained Llama-3.2-11B-based judge model outperforms both Llama-3.2-90B and Claude-3.5-Sonnet, while also achieving significant improvements on the hallucination dimension. These results are notable given that VLRB general draws from WildVision~\cite{lu2024wildvision} and VLFeedback~\cite{li2024vlfeedback}, which are datasets that capture authentic human preferences and diverse instruction-following scenarios from real-world VLM interactions. 
Our model's strong performance on this category indicates that our synthetic data generation pipeline effectively captures learning signals reflected by these real-world datasets.

\xhdr{MMRB General} The performance on MMRB General remains relatively flat compared to improvements observed in other dimensions. We hypothesize that this stems from the image diversity present in MMRB's benchmark composition. The general categories draw from two distinct datasets: NoCaps~\cite{agrawal2019nocaps} and VisitBench~\cite{bitton2023visit}. After 4 training iterations, we observe consistent performance gains on VisitBench (correctness 0.54 $\rightarrow$ 0.64), whereas NoCaps performance has either declined or plateaued (correctness 0.58 $\rightarrow$ 0.52). This divergence suggests a potential domain mismatch: NoCaps sources images from OpenImages~\cite{kuznetsova2020open}, which exhibits greater visual and object diversity for captioning tasks. The broader distribution of visual concepts in OpenImages likely challenges our model's ability to generalize beyond the training distribution. This finding underscores that both task diversity and image diversity are critical factors in synthetic data generation. While our approach successfully expands task diversity through the LLaVA-OneVision dataset~\cite{li2024llavaonevisioneasyvisualtask}, the results indicate that incorporating more diverse image sources could further enhance model robustness and generalization to visual domains such as those present in NoCaps.

\xhdr{MMRB Safety} The improvement on safety benchmarks is not as consistent as observed in other evaluation metrics. This limitation likely stems from the fact that our training data generation process did not explicitly prompt the model to produce biased or toxic content for creating synthetic pairs. The inconsistent safety performance emphasizes the critical importance of task alignment in synthetic data generation—safety improvements require training data that specifically addresses potential risks and harmful outputs. Since safety evaluation typically demands specialized guardrails and red-teaming approaches that extend beyond the scope of our current methodology, we leave comprehensive safety-focused synthetic data generation as an important direction for future work.

\section{Conclusion}

We present a self-improving framework for training vision-language model judges that eliminates the need for human preference annotations through strategic synthetic data generation and iterative refinement. Our approach demonstrates that a compact 11B parameter model can achieve competitive performance with significantly larger proprietary systems across multiple evaluation dimensions, particularly excelling in general instruction following, hallucination detection, and VQA tasks. Our analysis reveals clear pathways for further improvement through targeted data augmentation in specialized domains such as safety evaluation, while our core methodology remains applicable to emerging visual tasks where ground-truth annotations are scarce or unavailable.

\section*{Limitations}

\xhdr{Safety Evaluation Limitations} The modest improvements observed on safety benchmarks reflect a deliberate methodological boundary: our framework does not generate synthetic training data containing biased, toxic, or otherwise harmful content. Robust safety evaluation requires specialized infrastructure including red-teaming protocols, carefully designed adversarial examples, and nuanced understanding of subtle policy violations--components that fall outside the scope of our general-purpose self-improvement approach. Developing effective safety evaluation capabilities within self-improving frameworks remains an important open challenge, one that will require principled approaches for safely generating and learning from examples of harmful content without amplifying risks.

\xhdr{Task-Specific Performance Variation} Our synthetic data generation framework exhibits differential effectiveness across evaluation dimensions, reflecting the inherent complexity and diversity of multimodal judgment tasks. While substantial improvements emerge in general instruction following, hallucination detection, and visual question answering, other capabilities such as complex reasoning and nuanced preference judgments show more limited gains. This performance variance suggests potential benefits from mixture-of-experts architectures, where a routing mechanism could direct different judgment queries to specialized expert judges trained on domain-specific synthetic data. Exploring such multimodel approaches represents a promising direction for future work that could build upon our iterative self-training framework, extending the principle of self-generated supervision signals to specialized judge models tailored for different evaluation dimensions

\clearpage
\newpage
\bibliographystyle{assets/plainnat}
\bibliography{paper}

\clearpage
\newpage
\beginappendix

\section{Training example from different iterations}
\label{tab:reasoning_progression}
\begin{figure}[h!]
  \centering
  \includegraphics[width=0.4\textwidth]{images/training_iteration_ex.jpeg}
  \label{fig:your_label}
\end{figure}

\begin{table*}[h!]
\centering
\small
\begin{tabular}{>{\raggedright\arraybackslash}p{2.5cm}p{12cm}}
\toprule
\textbf{Component} & \textbf{Content} \\
\midrule
\textbf{Response A} & The image shows a blue double-decker bus driving down a street in a town or city. The bus is in the center of the image and has a large front window and a sign on the top that reads Brighton Open Top in yellow letters. The bus is driving away from the camera, and there are people standing on the sidewalk next to it. In the background, there are several buildings lining the street, including a row of shops and a few houses. The sky is overcast, and the overall atmosphere of the image is one of a quiet, everyday scene in a small town or city. \\
\midrule
\textbf{Response B} & The image shows a red double-decker bus driving down a street in a town, surrounded by tall buildings and parked cars. The bus has a red color and features a large front window, with the words "Brighton Open Top" written in yellow on the front. The bus appears to be in motion, with the front wheels turned to the left as it navigates the street. In the background, the buildings are tall and white, with many windows and balconies. Some of the buildings have signs or advertisements on them, but they are not clearly visible. The sky above is blue and cloudy, adding to the overall atmosphere of the scene. Overall, the image captures a typical urban scene, with the red bus being the main focus of attention. The surrounding buildings and parked cars add depth and context to the image, while the cloudy sky provides a sense of atmosphere and mood. \\
\midrule
\textbf{Iteration 1 Reasoning} & Both responses aim to describe the image, but they differ in their level of detail, accuracy, and relevance to the user's question. Assistant A's response provides a clear and concise description, accurately identifies the bus as blue, and notes people on the sidewalk. Assistant B's response provides more detailed description including the bus's color (red) and parked cars, and describes the surrounding buildings' height and signs. While both provide accurate information, Assistant B's response is more detailed and engaging, making it a better choice. \\
\midrule
\textbf{Iteration 2 Reasoning} & The two responses have similarities but notable differences. Assistant A's response is more detailed and descriptive, highlighting the red double-decker bus, tall buildings, and cloudy sky with simple, accessible language. Assistant B's response is more concise and lacks detail, only mentioning the blue double-decker bus and surrounding buildings without additional context. \\
\midrule
\textbf{Iteration 3 Reasoning} & The two responses differ significantly in accuracy, relevance, and depth. Assistant A's response is more accurate and relevant, correctly identifying the bus as red and describing features like the large front window and "Brighton Open Top" text. It provides detailed description of surrounding buildings and parked cars, adding context. Assistant B's response is less accurate, incorrectly identifying the bus as blue and failing to mention the red color. It lacks detailed description of surroundings, making it less informative. \\
\bottomrule
\end{tabular}
\caption{Comparison of assistant responses and judge reasoning across training iterations for the double-decker bus image example. Response A describes a blue bus while Response B describes a red bus. The judge's reasoning improves from iteration 1 (focusing on detail and engagement) to iteration 3 (explicitly identifying factual accuracy errors).}
\end{table*}

\newpage
\section{Judge Prompt}
\begin{verbatim}
Please act as an impartial judge and evaluate the quality of the 
responses provided by two AI assistants to the user question displayed 
below. You should choose the assistant that follows the user's 
instructions and answers the user's question better. Your evaluation 
should consider factors such as the helpfulness, relevance, accuracy, 
depth, creativity, and level of detail of their responses. Begin your 
evaluation by comparing the two responses and provide a short 
explanation. Avoid any position biases and ensure that the order in 
which the responses were presented does not influence your decision. 
Do not allow the length of the responses to influence your evaluation. 
Do not favor certain names of the assistants. Be as objective as 
possible. After providing your explanation, output your final verdict 
by strictly following this format: '[[A]]' if assistant A is better, 
'[[B]]' if assistant B is better.

[User Question]
{q}

[The Start of Assistant A's Answer]
{r1}
[The End of Assistant A's Answer]

[The Start of Assistant B's Answer]
{r2}
[The End of Assistant B's Answer]

YOU MUST STRICTLY FOLLOW THE FORMAT BELOW. Start with **EVALUATION** 
and make sure to add double brackets around the final answer.

**EVALUATION**: Provide a detailed comparison of both responses, 
analyzing their strengths and weaknesses based on the above factors. 
Be specific about why one response better serves the user's needs.

**FINAL ANSWER**: '[[A]]' if assistant A is better, '[[B]]' if 
assistant B is better.
\end{verbatim}

\section{Detail Alternation Prompts}
\label{app:alter_prompts}

\begin{verbatim}
INSTRUCTION="Below is a text (description, reasoning, explanation, etc.). 
Your task is to generate a version that incorporates alternative details 
while keeping original style and adhering to the following guidelines: 
1. Detail Modifications: Replace exactly TWO specific details (e.g., visual 
elements, facts, examples, reasoning steps, methods, premises) with 
alternative details, or change the perspective on TWO features. 2. Length, 
Structure and Style: Keep the overall length, structure, and style identical 
to the original. 3. Core Identity & Function: Ensure the main subject's 
identity and function remain unchanged. 4. Tone & Detail Level: Maintain the 
original tone and level of detail. 5. Style Mimicry: Mirror the exact writing 
style, vocabulary choices, sentence patterns, and linguistic quirks of the 
original text.; Below is a text (description, reasoning, explanation, etc.). 
Your task is to generate a reinterpreted version that incorporates ONE 
alternative detail while adhering to the following guidelines: 1. Detail 
Modifications: Replace exactly ONE specific detail (e.g., visual element, 
fact, example, reasoning step, method, premise) with alternative details, or 
change the perspective on ONE feature. 2. Length, Structure and Style: Keep 
the overall length, structure, and style identical to the original. 3. Core 
Identity & Function: Ensure the main subject's identity and function remain 
unchanged. 4. Tone & Detail Level: Maintain the original tone and level of 
detail. 5. Style Mimicry: Mirror the exact writing style, vocabulary choices, 
sentence patterns, and linguistic quirks of the original text.; Below is a 
text (description, reasoning, explanation, etc.). Your task is to generate a 
reinterpreted version that incorporates TWO alternative details while 
adhering to the following guidelines: 1. Detail Modifications: Replace 
exactly TWO specific details (e.g., visual elements, facts, examples, 
reasoning steps, methods, premises) with alternative details, or change the 
perspective on TWO features. 2. Length, Structure and Style: Keep the overall 
length, structure, and style identical to the original. 3. Core Identity & 
Function: Ensure the main subject's identity and function remain unchanged. 
4. Tone & Detail Level: Maintain the original tone and level of detail. 
5. Style Mimicry: Mirror the exact writing style, vocabulary choices, 
sentence patterns, and linguistic quirks of the original text.; Below is a 
text (description, reasoning, explanation, etc.). Your task is to generate a 
reinterpreted version that incorporates 1-2 alternative details while 
adhering to the following guidelines: 1. Detail Modifications: Replace 
exactly 1-2 specific details (e.g., visual elements, facts, examples, 
reasoning steps, methods, premises) with alternative details, or change the 
perspective on 1-2 features. 2. Length, Structure and Style: Keep the overall 
length, structure, and style identical to the original. 3. Core Identity & 
Function: Ensure the main subject's identity and function remain unchanged. 
4. Tone & Detail Level: Maintain the original tone and level of detail. 
5. You MUST make the required modifications at the beginning of the text. 
5. Style Mimicry: Mirror the exact writing style, vocabulary choices, 
sentence patterns, and linguistic quirks of the original text.; Below is a 
text (description, reasoning, explanation, etc.). Your task is to generate a 
reinterpreted version that incorporates 1-2 alternative details while 
adhering to the following guidelines: 1. Detail Modifications: Replace 
exactly 1-2 specific details (e.g., visual elements, facts, examples, 
reasoning steps, methods, premises) with alternative details, or change the 
perspective on 1-2 features. 2. Length, Structure and Style: Keep the overall 
length, structure, and style identical to the original. 4. Tone & Detail 
Level: Maintain the original tone and level of detail. 5. You MUST make the 
required modifications close to the END of the text. 5. Style Mimicry: Mirror 
the exact writing style, vocabulary choices, sentence patterns, and 
linguistic quirks of the original text.; Below is a text (description, 
reasoning, explanation, etc.). Your task is to generate a reinterpreted 
version that incorporates 1-2 alternative details while adhering to the 
following guidelines: 1. Detail Modifications: Replace exactly 1-2 specific 
details (e.g., visual elements, facts, examples, reasoning steps, methods, 
premises) with alternative details, or change the perspective on 1-2 
features. 2. Length, Structure and Style: Keep the overall length, structure, 
and style identical to the original. 3. Core Identity & Function: Ensure the 
main subject's identity and function remain unchanged. 4. Tone & Detail 
Level: Maintain the original tone and level of detail. 5. You MUST make the 
required modifications close to the MIDDLE of the text, instead of at the 
beginning or the end. 5. Style Mimicry: Mirror the exact writing style, 
vocabulary choices, sentence patterns, and linguistic quirks of the original 
text."
\end{verbatim}

\section{Generation Meta Prompt for Captions}
\label{app:generation_prompt_caption}

\begin{verbatim}
    {
    "llava_image_description_prompts": {
        "concise": [
            "Describe the image concisely.",
            "Provide a brief description of the given image.",
            "Offer a succinct explanation of the picture presented.",
            "Summarize the visual content of the image.",
            "Give a short and clear explanation of the subsequent image.",
            "Share a concise interpretation of the image provided.",
            "Present a compact description of the photo's key features.",
            "Relay a brief, clear account of the picture shown.",
            "Render a clear and concise summary of the photo.",
            "Write a terse but informative summary of the picture.",
            "Create a compact narrative representing the image presented."
        ],
        "detailed": [
            "Describe the following image in detail",
            "Provide a detailed description of the given image",
            "Give an elaborate explanation of the image you see",
            "Share a comprehensive rundown of the presented image",
            "Offer a thorough analysis of the image",
            "Explain the various aspects of the image before you",
            "Clarify the contents of the displayed image with great detail",
            "Characterize the image using a well-detailed description",
            "Break down the elements of the image in a detailed manner",
            "Walk through the important details of the image",
            "Portray the image with a rich, descriptive narrative",
            "Narrate the contents of the image with precision",
            "Analyze the image in a comprehensive and detailed manner",
            "Illustrate the image through a descriptive explanation",
            "Examine the image closely and share its details",
            "Write an exhaustive depiction of the given image"
        ]
    }
}
\end{verbatim}

\begin{verbatim}
{
    "caption_styles": {
        "concise":{
            "chatgpt": "The image shows a golden retriever sitting 
in a sunny park with green grass and trees in the background. 
The dog appears happy and is looking directly at the camera 
with its tongue out. There are some fallen leaves scattered 
around, suggesting it might be autumn. The lighting is natural 
and warm, creating a pleasant, cheerful atmosphere.",
            "claude": "A golden retriever sits attentively in a 
park setting, its mouth open in what appears to be a content 
pant. The warm lighting and scattered autumn leaves create a 
peaceful scene, with the dog positioned centrally against a 
backdrop of grass and mature trees. The composition suggests a 
casual outdoor moment captured during a walk or play session.",
            "qwen": "Golden retriever dog sitting on grass in park 
environment. Autumn season indicated by fallen leaves on ground. 
Natural daylight illumination. Dog exhibits typical friendly 
expression with open mouth. Background contains trees and 
vegetation. Image quality appears high resolution with good 
color saturation.",
            "llama": "This image depicts a golden retriever in an 
outdoor setting. The dog is positioned on grass with trees 
visible in the background. The scene appears to be during 
daytime with natural lighting. The dog's posture suggests it 
is alert and engaged. The presence of fallen leaves indicates 
the photograph was likely taken during autumn months.",
            "gemini": "A beautiful golden retriever enjoys a moment 
in the park! The friendly pup is sitting on the grass, 
surrounded by the warm colors of fall. With its tongue happily 
hanging out, this good boy seems to be having a wonderful time 
outdoors. The natural lighting and peaceful setting make for a 
heartwarming scene that captures the joy of spending time with 
our four-legged friends."
        }
    }
}
\end{verbatim}

\section{Hyperparameters}
\label{app:hyperparameters}

In data generation, we use a temperature of 0.6, top\_k=50, top\_p=0.9.

In data sampling for majority voting (Section~\ref{sec:method}), we set N = 16 to generate 
16 candidate responses per closed-ended question. We experimented with different 
majority thresholds and found that requiring at least 5 identical responses 
(out of 16) provides the best balance between data quality and quantity. Lower 
thresholds (e.g., 3) produce more training data but include less reliable 
majorities, while higher thresholds (e.g., 8) overly restrict the dataset. 
We selected the threshold of 5 after empirically comparing thresholds of 3 and 5: 
both achieved similar final performance, but the threshold of 5 required less 
training data, suggesting better sample efficiency and filtering quality. This 
threshold effectively filters out cases with greater response diversity where 
the model's uncertainty is higher, retaining only examples where the model 
shows reasonable consensus.

In training, we use the default training setup in llama-cookbook which has a learning rate of 1e-5, we train for 5 epochs.

The data generation and model training were conducted on H100 GPUs, with approximately 400 GPUs hours. This represents a cost-efficient alternative to human 
annotation (~\$500K-1M for equivalent data) or GPT-4 distillation 
(~\$5K-10K in API costs).

\newpage
\section{Correctness Filter Negative Example}
\label{tab:counter_example}

\begin{figure}[h]
  \centering
  \includegraphics[width=0.2\textwidth]{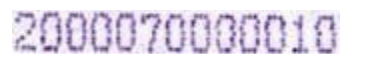}
  \label{fig:your_label}
\end{figure}

\begin{table*}[h!]
\centering
\small
\begin{tabular}{>{\raggedright\arraybackslash}p{2.5cm}p{12cm}}
\toprule
\textbf{Component} & \textbf{Content} \\
\midrule
\textbf{Response A} & The image contains a series of numbers, specifically ``20000700000010''. \\
\midrule
\textbf{Response B} & The image contains the text ``2000070000010'' in a purple, pixelated font. The text is centered at the top of the image and spans the entire width. It appears to be a numerical code or identifier, but without more context, it's difficult to determine its meaning or significance. \\
\midrule
\textbf{Evaluation} & Both responses attempt to decipher the content of the image, but they differ in their approach and level of detail.

Assistant A's response is concise and straightforward, simply stating the numbers visible in the image. However, it lacks context and fails to provide any additional information about the meaning or significance of these numbers.

Assistant B's response provides more detailed observations, including information about the font style, color, and layout of the text. It also acknowledges the limitations of interpretation without additional context.

\textbf{FINAL ANSWER}: [[B]] \\
\bottomrule
\end{tabular}
\caption{Example of judge getting the correct answer but providing incorrect reasoning. Even though the final answer is correct, the reasoning failed to point out that the number in Response A has an extra 0 and it is a wrong answer.}
\label{tab:counter_example}
\end{table*}

While our filtering mechanism successfully removes many incorrect judgments, 
we acknowledge it is not perfect. This section illustrates a case where correctness-
based filtering can retain problematic examples: the judge selects the correct 
answer (Response B) but fails to identify the specific transcription error in 
Response A (an extra digit "20000700000010" vs. "2000070000010"). Instead, the 
reasoning focuses on superficial features—Response B's additional descriptive 
details and contextual observations—rather than the factual discrepancy that 
should determine the judgment.

This example suggests why filtering based on ground-truth correctness alone may 
be insufficient--even when the final judgment is correct, the reasoning may 
prioritize stylistic features over factual accuracy. This observation partially 
motivates our majority voting approach (Section~\ref{sec:analysis_reasoning}), which filters training 
data based on the model's own consistency rather than relying solely on external 
correctness labels. We hypothesize that by requiring the judge to correctly 
identify preferences across multiple synthetically generated pairs with diverse 
error types, majority voting may implicitly filter for more robust reasoning 
patterns rather than spurious correlations with superficial features. A judgment 
that succeeds consistently across varied synthetic contrasts could provide 
stronger evidence of genuine understanding than a single correct answer that 
might result from incidental heuristics.

We include this example not as definitive evidence of a fundamental flaw, but 
as an illustrative case that helps explain our design rationale. While we cannot 
conclusively verify that majority voting eliminates all imperfect reasoning, 
the empirical advantages shown in Section~\ref{sec:analysis_reasoning} suggest that consistency-based 
filtering may provide more robust supervision than correctness checking alone 
for learning generalizable judgment criteria.


\end{document}